\documentclass{article}

\usepackage[final]{neurips_2021}

\usepackage{graphicx}

\usepackage{amsmath}
\usepackage{amsthm}
\usepackage{algorithm}
\usepackage{algorithmic}
\usepackage{booktabs}
\usepackage{bbding}
\usepackage{multirow}
\usepackage{color}
\usepackage{amsfonts}
\usepackage{bm}
\usepackage{CJKutf8}
\usepackage{array}
\usepackage{etoolbox}
\usepackage[bottom]{footmisc}
\usepackage{natbib}
\newcolumntype{L}[1]{>{\raggedright\let\newline\\\arraybackslash\hspace{0pt}}m{#1}}
\newcolumntype{C}[1]{>{\centering\let\newline\\\arraybackslash\hspace{0pt}}m{#1}}
\newcolumntype{R}[1]{>{\raggedleft\let\newline\\\arraybackslash\hspace{0pt}}m{#1}}

\usepackage[utf8]{inputenc} 
\usepackage[T1]{fontenc}    
\usepackage{hyperref}       
\usepackage{url}            
\usepackage{booktabs}       
\usepackage{amsfonts}       
\usepackage{nicefrac}       
\usepackage{microtype}      
\usepackage{xcolor}         

\usepackage{subcaption}
\usepackage{cleveref}

\title{Unsupervised Domain Adaptation with Adapter}

\author{%
  Rongsheng Zhang$^{1}$\thanks{\quad Equal contribution.}~~\thanks{\quad Corresponding author.}~, Yinhe Zheng$^{2,3}$\footnotemark[1]~, Xiaoxi Mao$^1$, Minlie Huang$^2$ \\
  $^1$ {Fuxi AI Lab, NetEase Inc., Hangzhou, China} \\
  $^2$ {Department of Computer Science and Technology, Institute for Artifical Intelligence, State Key} \\
  {Lab of Intelligent Technology and Systems, Beijing National Research Center for} \\
  {Information Science and Technology, Tsinghua University, Beijing, China.} \\
  $^3$ {Alibaba Group, Beijing, China} \\
  \texttt{zhangrongsheng@corp.netease.com, zhengyinhe.zyh@alibaba-inc.com} 
}

\begin{document}

\maketitle

\begin{abstract}
  Unsupervised domain adaptation (UDA) with pre-trained language models (PrLM) has achieved promising results since these pre-trained models embed generic knowledge learned from various domains. However, fine-tuning all the parameters of the PrLM on a small domain-specific corpus distort the learned generic knowledge, and it is also expensive to deployment a whole fine-tuned PrLM for each domain. This paper explores an adapter-based fine-tuning approach for unsupervised domain adaptation. Specifically, several trainable adapter modules are inserted in a PrLM, and the embedded generic knowledge is preserved by fixing the parameters of the original PrLM at fine-tuning. A domain-fusion scheme is introduced to train these adapters using a mix-domain corpus to better capture transferable features. Elaborated experiments on two benchmark datasets are carried out, and the results demonstrate that our approach is effective with different tasks, dataset sizes, and domain similarities.
\end{abstract}

\section{Introduction}
Unsupervised domain adaption (UDA) is an essential task in the realm of deep learning since it mitigates the expensive burden of manual annotation by focusing on cheap unlabeled data from target domains \citep{ramponi2020neural}. Among all existing approaches for UDA, pre-trained language model (PrLM) based approaches become the de-facto standard \citep{gururangan-etal-2020-dont, ben2020perl, yu2021adaptsum,  karouzos2021udalm} since these PrLMs are equipped with generic knowledge learned from large corpora \citep{howard2018universal} and lead to promising results.

\begin{figure*}[!t]
\setlength{\belowcaptionskip}{-0.3cm}
\centering
\includegraphics[width=340px]{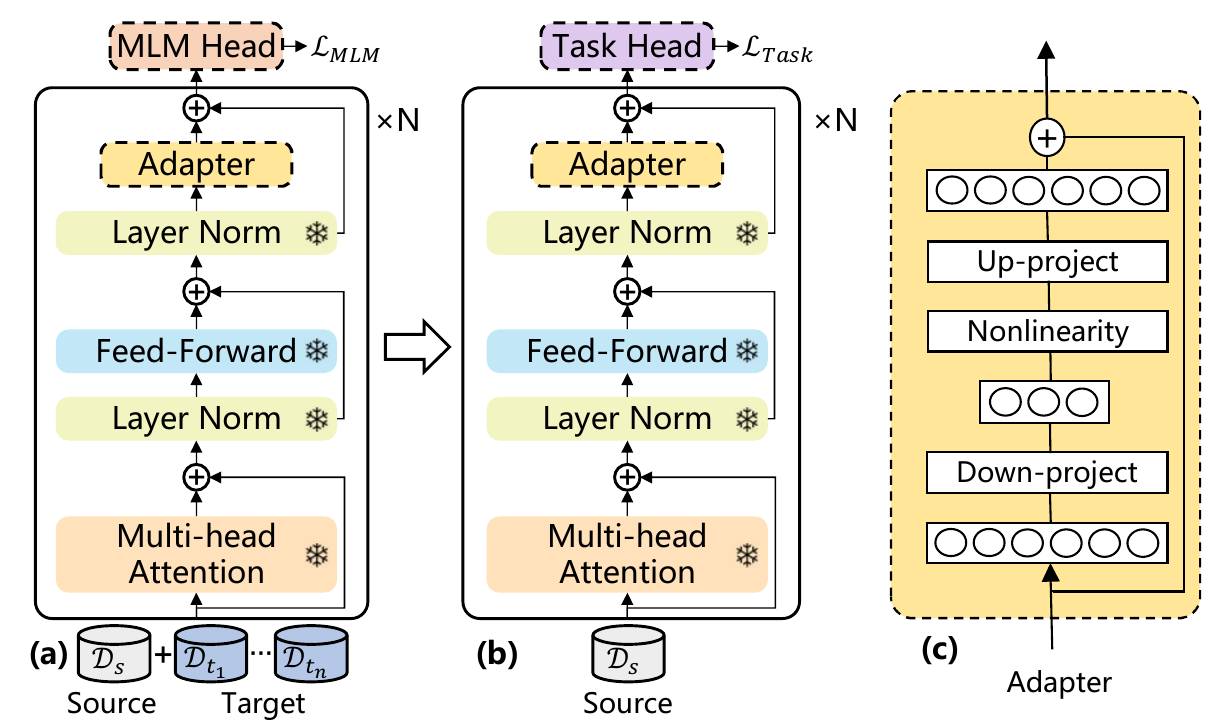}
\caption{UDA with the adapter.
a) Domain-fusion training: train the adapter and MLM head using the MLM loss on a mixed corpus covering both source and target domains;
b) Task fine-tuning: fine-tune the adapter and a task-specified head using the task specified loss on source domain samples.
Note that all the parameters for the underlying pre-trained transformer model (i.e., sub-modules marked with {\small \SnowflakeChevron}) are fixed. c) The architecture of the adapter module.}
\label{fig:uda_with_adapter}
\end{figure*}

The primary focuses of UDA methods are to capture the transferable features for the target domain while reserving the knowledge learned from the source domain \citep{blitzer2006domain, pan2010cross}. However, most existing pre-training-based UDA approaches are carried out by fine-tuning the entire set of model parameters on domain-specific corpora \citep{gururangan-etal-2020-dont, yu2021adaptsum,  karouzos2021udalm}, which are usually of limited sizes. Such a setting may easily drift the PrLM to a specified domain and distort the generic knowledge embedded in the original PrLM weights \citep{pfeiffer2020adapterfusion, he2021effectiveness}. This hinders the model from capturing transferable features between different domains and leads to sub-optimal performance for UDA tasks \citep{karouzos2021udalm}. Moreover, it is also expensive to fine-tune and deploy a large model for every single domain \citep{houlsby2019parameter}.

We observe that the intuition of preserving learned knowledge coincides with recently developed adapter-based tuning methods \citep{houlsby2019parameter, rebuffi2017learning}, in which several trainable adapter modules are introduced between layers of a pre-trained language model (LM) while parameters from the original LM are fixed. This setting helps preserve the knowledge embedded in the PrLM and alleviates the distortion of features for different domains since the original PrLM remains intact \citep{he2021effectiveness,pfeiffer2020adapterfusion, pfeiffer2020adapterhub,  houlsby2019parameter}. However, few studies are performed to extend this effective method to tackle UDA tasks.

In this paper, we explore to introduce adapter modules in pre-training-based UDA approaches. Specifically, several bottle-necked adapter modules are inserted in a transformer-based PrLM \citep{vaswani2017attention}. These adapters are learned following a two-step process: 1) The \textbf{domain-fusion training} step trains adapters with the Masked-Language-Model (MLM) loss \citep{devlin2019bert} on a mixed corpus containing data from both the source and all the target domains. This step facilitates the capture and fusion of transferable knowledge between different domains; 2) The \textbf{task fine-tuning} step fine-tunes adapters with the task-specific loss on the source domain corpus. Note that parameters of the underlying pre-trained LM are fixed throughout the two learning processes. This helps prevent the drifting of learned generic knowledge and facilitates more effective domain knowledge transferring \citep{pfeiffer2020adapterhub}. In the testing phase, we apply the resulted model to data sampled from the target domain. The results on two benchmark datasets indicate that our method outperforms competitive baselines and is effective in improving the performance of downstream UDA tasks \footnote{Our code is available in Appendix \ref{sec:link}}.

Our contributions can be summarized as:

1. We apply adapter modules in the pre-training-based UDA approaches. Specifically, trainable adapters are introduced in a PrLM, and a two-step process is introduced to facilitate the learning of these adapters.

2. Elaborated experiments on two benchmark datasets show that our approach outperforms competitive baselines and is more effective to improve the performance of downstream UDA tasks.

\section{Related Work}
\textbf{Unsupervised Domain Adaption:}
Existing UDA approaches can be generally classified into two categories: 1) The \emph{model-based} methods target at augmenting the feature spaces \citep{glorot2011domain, chen2012marginalized, ziser2019task, ben2020perl}, designing new losses \citep{ganin2015unsupervised, ganin2016domain} or refining model structures \citep{bousmalis2016domain}; 2) The \emph{data-based} methods aim to utilize pseudo-labels \citep{ruder2018strong, lim2020semi} and develop better data selection schemes \citep{han2019unsupervised, ma2019domain}. Some works also try to tackle UDA tasks utilizing large PrLM \citep{li2019semi,gururangan-etal-2020-dont, yu2021adaptsum, karouzos2021udalm}, which are becoming the de-facto standard for various NLP tasks. Although promising results are reported, fine-tuning the whole model on a small amount of domain-specific data may distort underlying PrLM and lead to sub-optimal performances.

\textbf{Adapters}:
In NLP studies, adapter modules are primarily used for parameter-efficient fine-tuning of large PrLMs \citep{lauscher2020common, wang2020k, lin2021adapter, poth2021pre, han2021robust,mahabadi2021parameter}. The most similar works comparing to our study are the models for zero-shot cross-lingual transfer tasks \citep{pfeiffer2020mad, vidoni2020orthogonal}. However, these models aim to separate language-specific knowledge using adapters, while our UDA task tries to capture common and transferable features across different domains.

\section{Method}\label{sec:method}
\subsection{Task Formulation}
The UDA task investigated in this study aims to improve the model performance with the help of unlabeled data. Specifically, the training data consists of two parts: 1) labeled dataset $\mathcal{D}_s = \{(x^s_j, y_j)\}$ collected from a single source domain $\mathcal{S}$; 2) $n$ unlabeled datasets $\mathcal{D}_{t_i} = \{x^{t_i}_j\}$ $(i=1,...,n)$ collected from $n$ target domains $\mathcal{T}_i$ $(i=1,...,n)$. $x^s_j$ and $x^{t_i}_j$ represents data drawn from $\mathcal{S}$ and $\mathcal{T}_i$, respectively, and $y_j \in \mathcal{Y}$ is the label associated with $x^{t_i}_j$, where $\mathcal{Y}$ is the label space. In the testing phase, we examine the effect of UDA on the labeled dataset $\bar{\mathcal{D}}_{t_i} = \{(\bar{x}^{t_i}_j, \bar{y}_j)\}$ collected for each target domain $\mathcal{T}_{i}$, in which $\bar{y}_j \in \mathcal{Y}$.

\subsection{Adapters Architecture}
Figure \ref{fig:uda_with_adapter} shows an overview of our adapter-based UDA approach. Specifically, a transformer model is first initialized using a set of pre-trained weights, and a trainable adapter module is inserted into each transformer layer. Here we apply a variant of the efficient lightweight-adapter \citep{pfeiffer2020adapterfusion} that only adds one bottle-necked MLP after the feed-forward sublayer of each transformer block. The bottle-necked MLP first projects the $H$-dimensional input representations into a smaller dimension $m$ ($m<H$), then applies a nonlinearity (GELU \citep{hendrycks2016gaussian} in our case), and finally projects back to $H$ dimensions. A residual connection is applied across the adapter.

\subsection{Two-step Adaption}\label{sec:two_step}

In the training process, we fix the parameters initialized from the PrLM and propose to learn the parameters of the randomly initialized adapter modules using a two-step process:

The first step is {\bf domain-fusion training}, in which we mix the training instances from both the source and target domains, $\mathcal{D}_s \bigcup \mathcal{D}_{t_i} ... \bigcup  \mathcal{D}_{t_n}$, and train the adapters with the MLM loss $\mathcal{L}_{MLM}$ on these instances. This setting enables the adapter to capture transferable features among all the domains, which enriches the knowledge embedded in the pre-trained model.

The second step, i.e., {\bf task fine-tuning}, learns a task head using the task-specific loss $\mathcal{L}_{Task}$ (for example, the cross-entropy loss for classification tasks) on the labeled dataset $\mathcal{D}_s$. In this step, we also learn the adapters' parameters to allocate more modeling capacity to fit the task distribution. Note that different from the adaptive pre-training method \citep{gururangan-etal-2020-dont} that directly learns the task head on labeled target domain data, our study follows the UDA setting that learns the task head on the source domain and tests it on the target domain. The transferable features captured in adapters help generalize the learned task knowledge and improve the performance of the resulting model on target domains.

\section{Experiments}
\subsection{Dataset}
Our UDA approach is evaluated on two benchmark datasets with different tasks:

1. \textbf{SDA:}
\underline{S}entiment \underline{D}omain \underline{A}daptation dataset \citep{blitzer2007biographies} that contains Amazon product reviews for four different product types (i.e., domains): \underline{B}ooks ($\mathcal{B}$), \underline{D}vds ($\mathcal{D}$), \underline{E}lectronics ($\mathcal{E}$) and \underline{K}itchen appliances ($\mathcal{K}$). 2.0K reviews with binary sentiment labels are available for each domain and we split these data into Train and Dev set with a ratio of 8:2. Four adaption schemes are attempted by regarding each of these four domains as the source domain ($src$) and the rest domains as the target domain ($tgt$). Test set of each domain contains 1.6K labeled reviews with binary sentiment labels.

2. \textbf{XNLI:}
Cross-lingual \underline{N}atural \underline{L}anguage \underline{I}nference dataset \citep{conneau2018xnli} that focuses on the NLI three-way classification task. This dataset involves texts with 15 languages, and the Train, Dev, and Test set of each language contain 392.70K, 2.49K, and 5.01K samples, respectively. In this study, we regard each language as a domain and only use English as the source domain.

\subsection{Implementation Details}

\textbf{Model and Training:}
For experiments on the SDA and XNLI dataset, the pre-trained $\text{RoBERTa}_\text{base}$ \citep{liu2019roberta} and $\text{XLM-R}_\text{base}$ \citep{conneau-etal-2020-unsupervised} model is used as the initialization PrLM, respectively, and the adapter size $m$ is set to 128, and 256, respectively. The Adam optimizer \citep{kingma2014adam} is used with a learning rate of 5e-5. The domain-fusion training process lasts for 10 epochs and the best models are selected based on the Dev set performance. See Appendix \ref{sec:training_details} for more training details.

\paragraph{Baselines:}
Three baselines are tested in our study. The first two baselines do not utilize the adapter module: 1) \textbf{Full-FT} \underline{F}ine-\underline{T}unes all transformer parameters using $\mathcal{L}_{Task}$ on $\mathcal{D}_S$. The domain-fusion training is not performed in this baseline; 2) \textbf{Full-TSA} adjusts the work of \citep{gururangan-etal-2020-dont} to UDA tasks, i.e., the \underline{T}wo-\underline{S}tep \underline{A}daption process introduced in Section \ref{sec:two_step} is applied on all transformer parameters. The third baseline in our study, 3) \textbf{Ada-FT}, incorporates the adapter module but do not perform the domain-fusion training process. We denote our approach as \textbf{Ada-TSA}.

\begin{table*}[!tbh]
\small
\setlength{\belowcaptionskip}{-0.3cm}
\centering
\setlength\tabcolsep{7pt}
\begin{tabular}{c|ccc|c}
\toprule
$src \rightarrow tgt$ & Full-FT & Full-TSA & Ada-FT & Ada-TSA \\
\midrule 
$\mathcal{B} \rightarrow \mathcal{D}$ &  92.80 \small $\pm$ 0.30  & 92.80 \small $\pm$ 0.40 & 92.46 \small$\pm$ 0.42 & \textbf{92.87} \small$\pm$ 0.27 \\
$\mathcal{B} \rightarrow \mathcal{E}$ &  92.01 \small $\pm$ 0.67  & 91.26$^\dag$\small $\pm$ 1.15 & 91.98$^\dag$\small$\pm$ 0.54 & \textbf{92.84} \small$\pm$ 0.43 \\
$\mathcal{B} \rightarrow \mathcal{K}$ &  93.70$^\dag$\small $\pm$ 0.09  & \textbf{94.36} \small $\pm$ 0.59 & 92.91$^\dag$\small$\pm$ 0.88 & 93.98 \small$\pm$ 0.16 \\
\midrule 
$\mathcal{D} \rightarrow \mathcal{B}$ &  92.25$^\dag$\small $\pm$ 0.81  &  \textbf{93.74} \small $\pm$ 0.66 & 92.51 \small$\pm$ 1.65 & 93.30 \small$\pm$ 0.28 \\
$\mathcal{D} \rightarrow \mathcal{E}$ &  91.46$^\ddag$\small $\pm$ 0.40  & 93.17 \small $\pm$ 0.54 & 91.50$^\ddag$\small$\pm$ 0.47 & \textbf{93.72} \small$\pm$ 0.17 \\
$\mathcal{D} \rightarrow \mathcal{K}$ &  92.64$^\ddag$\small $\pm$ 0.68  & 94.17 \small $\pm$ 0.20 & 93.41 \small$\pm$ 0.86 & \textbf{94.39} \small$\pm$ 0.32 \\
\midrule
$\mathcal{E} \rightarrow \mathcal{B}$ &  91.44$^\ddag$\small $\pm$ 0.26  & 91.41$^\ddag$\small $\pm$ 0.24 & 90.77$^\ddag$\small$\pm$ 0.31 & \textbf{92.25} \small$\pm$ 0.29 \\
$\mathcal{E} \rightarrow \mathcal{D}$ &  91.34$^\ddag$\small $\pm$ 0.12  & 91.66 \small $\pm$ 0.29 & 91.57 \small$\pm$ 0.58 & \textbf{92.15} \small$\pm$ 0.33 \\
$\mathcal{E} \rightarrow \mathcal{K}$ &  93.48$^\ddag$\small $\pm$ 0.19  & 94.10 \small $\pm$ 0.16 & \textbf{94.32} \small$\pm$ 0.36 & 94.14 \small$\pm$ 0.21 \\
\midrule
$\mathcal{K} \rightarrow \mathcal{B}$ &  90.77$^\ddag$\small $\pm$ 0.21  & 91.89$^\ddag$\small $\pm$ 0.14 & 90.54$^\ddag$\small$\pm$ 0.45 & \textbf{93.20} \small$\pm$ 0.17 \\
$\mathcal{K} \rightarrow \mathcal{D}$ &  90.85$^\ddag$\small $\pm$ 0.31  & 92.54$^\ddag$\small $\pm$ 0.08 & 91.01$^\ddag$\small$\pm$ 0.51 & \textbf{93.26} \small$\pm$ 0.07 \\
$\mathcal{K} \rightarrow \mathcal{E}$ &  92.54$^\ddag$\small $\pm$ 0.22  & \textbf{93.65} \small $\pm$ 0.13 & 93.09 \small$\pm$ 0.81 & 93.63 \small$\pm$ 0.16 \\
\midrule
Avg. &  92.10$^\ddag$  & 92.89$^\dag$  & 92.17$^\ddag$ & \textbf{93.31} \\
\bottomrule
\end{tabular}
\caption{Sentiment classification accuracy and standard deviations of five independent runs on SDA. $\dag$ and $\ddag$ indicates significant difference ($t$-test) between our approach and the baseline with $p$-value \textless~0.05 and 0.01, respectively.}
\label{tab:sda_result}
\end{table*}

\begin{table*}[!tbh]
\small
\setlength{\belowcaptionskip}{-0.3cm}
\centering
\setlength\tabcolsep{7pt}
\begin{tabular}{c|ccc|c}
\toprule
$tgt$ & Full-FT & Full-TSA & Ada-FT & Ada-TSA \\
\midrule
en &  83.16$^\ddag$\small $\pm$ 0.30  & 83.14$^\ddag$\small $\pm$ 0.29 & 83.98 \small$\pm$ 0.39 & \textbf{84.57} \small$\pm$ 0.41 \\
fr &  76.57$^\ddag$\small $\pm$ 0.52  & 76.22$^\ddag$\small $\pm$ 0.21 & 77.11 \small$\pm$ 0.72 & \textbf{77.91} \small$\pm$ 0.51 \\
es &  77.24$^\dag$\small $\pm$ 0.39  & 77.87 \small $\pm$ 0.32 & 78.11 \small$\pm$ 0.37 & \textbf{78.17} \small$\pm$ 0.58 \\
de &  75.06$^\dag$\small $\pm$ 0.32  &  75.96 \small $\pm$ 0.44 & 74.90$^\dag$\small$\pm$ 0.37 & \textbf{76.02} \small$\pm$ 0.57 \\
el &  73.61$^\ddag$\small $\pm$ 0.25  & 74.63$^\ddag$\small $\pm$ 0.29 & 74.40 \small$\pm$ 0.72 & \textbf{75.21} \small$\pm$ 0.34 \\
bg &  76.25$^\ddag$\small $\pm$ 0.13  & 76.86 \small $\pm$ 0.42 & 77.30 \small$\pm$ 0.42 & \textbf{77.43} \small$\pm$ 0.46 \\
ru &  73.89$^\ddag$\small $\pm$ 0.21  & 74.45$^\dag$\small $\pm$ 0.39 & 75.11 \small$\pm$ 0.30 & \textbf{75.39} \small$\pm$ 0.47 \\
tr &  70.57$^\ddag$\small $\pm$ 0.21  & 71.13$^\dag$\small $\pm$ 0.13 & 71.96 \small$\pm$ 0.39 & \textbf{71.97} \small$\pm$ 0.56 \\
ar &  69.64$^\ddag$\small $\pm$ 0.17  & 70.22$^\ddag$\small $\pm$ 0.46 & 70.49$^\dag$\small$\pm$ 0.51 & \textbf{71.31} \small$\pm$ 0.37 \\
vi &  73.37 \small $\pm$ 0.50  & 73.59 \small $\pm$ 0.33 & \textbf{74.15} \small$\pm$ 0.53 & 73.86 \small$\pm$ 0.49 \\
th &  70.33 \small $\pm$ 0.55  & 71.20$^\ddag$\small $\pm$ 0.29 & \textbf{71.29}$^\dag$\small$\pm$ 0.79 & 69.99 \small$\pm$ 0.62 \\
zh &  72.18 \small $\pm$ 0.20  & \textbf{74.47}$^\ddag$\small $\pm$ 0.29 & 72.48 \small$\pm$ 0.58 & 72.98 \small$\pm$ 0.51 \\
hi &  68.74 \small $\pm$ 0.52  & 69.36 \small $\pm$ 0.52 & \textbf{69.36} \small$\pm$ 0.37 & 68.69 \small$\pm$ 0.62 \\
sw &  60.98$^\ddag$\small $\pm$ 0.34  & \textbf{65.45} \small $\pm$ 0.42 & 63.31$^\ddag$\small$\pm$ 0.21 & 64.88 \small$\pm$ 0.54 \\
ur &  64.34 \small $\pm$ 0.22  & 62.96$^\dag$\small $\pm$ 0.63 & \textbf{64.95} \small$\pm$ 0.52 & 64.15 \small$\pm$ 0.54 \\
\midrule
Avg. &  71.62$^\ddag$  & 72.45  & 72.49 & \textbf{72.71} \\
\bottomrule
\end{tabular}
\caption{NLI accuracy and standard deviations of five independent runs on XNLI dataset. $\dag$ and $\ddag$ carry the same means as in Table~\ref{tab:sda_result}.}
\label{tab:xnli_result}
\end{table*}


                                                                         





\subsection{Results and Discussion}
\begin{figure*}[ht]
\centering
\setlength{\belowcaptionskip}{-0.5cm}
\includegraphics[width=260px]{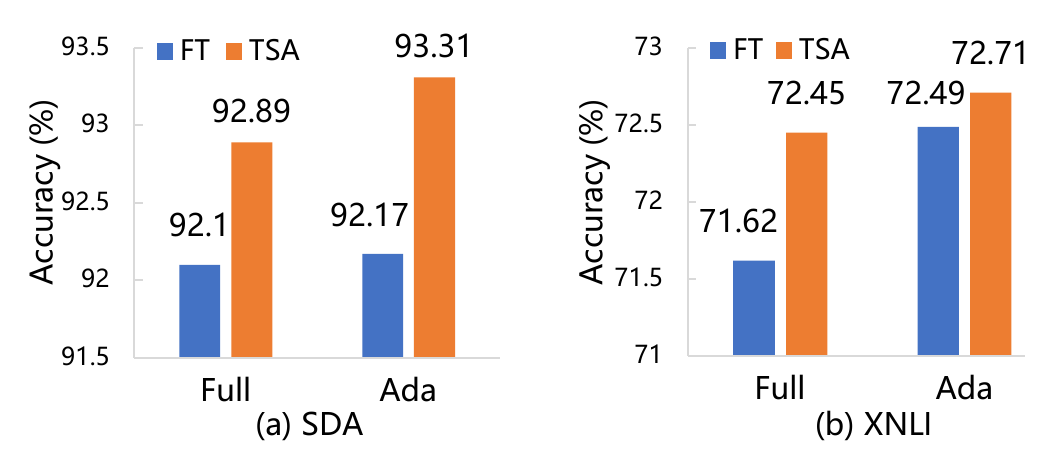}
\caption{Averaged performance on both datasets.}
\label{fig:avg_results}
\end{figure*}

\textbf{UDA Results:}
The results of our approach and baselines are summarized in Table \ref{tab:sda_result} (for the sentiment classification task on SDA) and Table \ref{tab:xnli_result} (for the NLI task on XNLI). All results are averaged over five runs with different random seeds. Our approach achieves the best results in most adaptation settings. Moreover, Figure \ref{fig:avg_results} shows the averaged performance of each model across all adaption settings. It can be seen that applying the adapter module and using the two-step adaption approach helps to improve the UDA performance. Specifically, for the averaged performance, our method yields an absolute improvement of 1.21\% and 1.09\% over the Full-FT baseline on the SDA and XNLI dataset, respectively. Results on more datasets in Appendix \ref{sec:sup_exp} also support this conclusion.

\paragraph{Adapter for Different Data Sizes:}
Since the SDA and XNLI datasets have different sizes, the results reported in Table \ref{tab:sda_result} and \ref{tab:xnli_result} also reveal the effect of data size on UDA performances:

1) When $\mathcal{D}_S$ is small, the adapter module is less effective without the domain-fusion step. Specifically, if we do not apply the domain-fusion training process, the adapter module only improves the averaged performance from 92.10\% (Full-FT) to 92.17\% (Ada-FT) on the relatively small dataset SDA (i.e., an absolute improvement of 0.07\%). However, on the XNLI dataset, which contains more data, the performance gain brought by the adapter module is much larger, i.e., from 71.62\% (Full-FT) to 72.49\% (Ada-FT). This may be because a small $\mathcal{D}_S$ is not enough to properly learn the adapter module from random initialization.

2) Our two-step adaption process improves the effectiveness of the adapter on small $\mathcal{D}_S$. Specifically, when the two-step adaption process is applied, the adapter module becomes more effective on the SDA dataset that an absolute improvement of 0.42\% (i.e., from 92.89\% of Full-TSA to 93.31\% of Ada-TSA) is observed. That is, our two-step adaption process upgrades the effectiveness of the adapter about 6 times on SDA.

\begin{figure}[htbp]
\setlength{\belowcaptionskip}{-0.1cm}
  \centering
    \begin{subfigure}{0.45\textwidth}
      \centering   
      \includegraphics[width=150px]{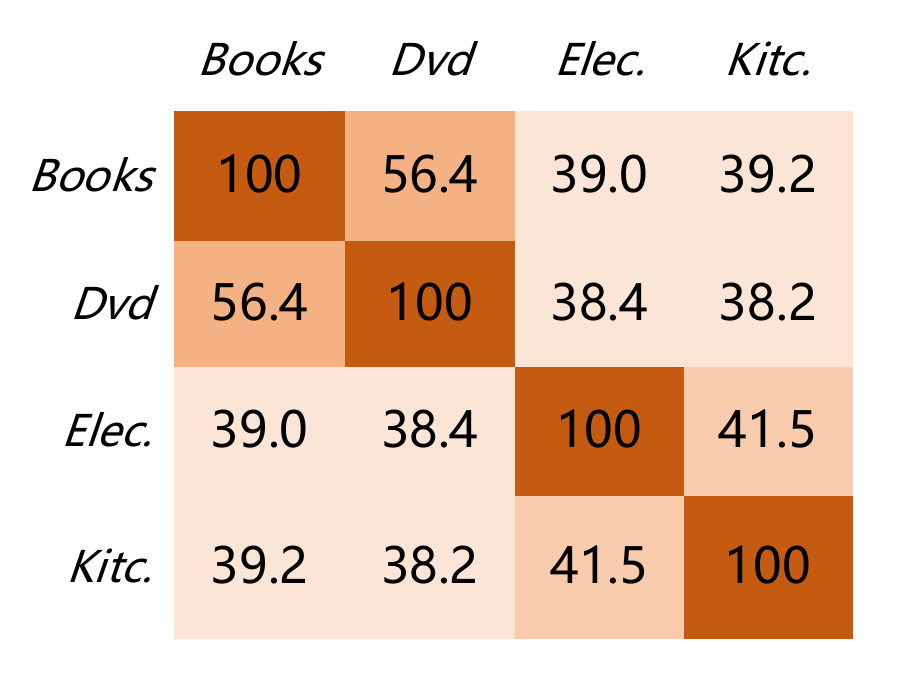}
        \caption{SDA}
        \label{fig:sda_sim}
    \end{subfigure}   
    \begin{subfigure}{0.45\textwidth}
      \centering   
      \includegraphics[width=135px]{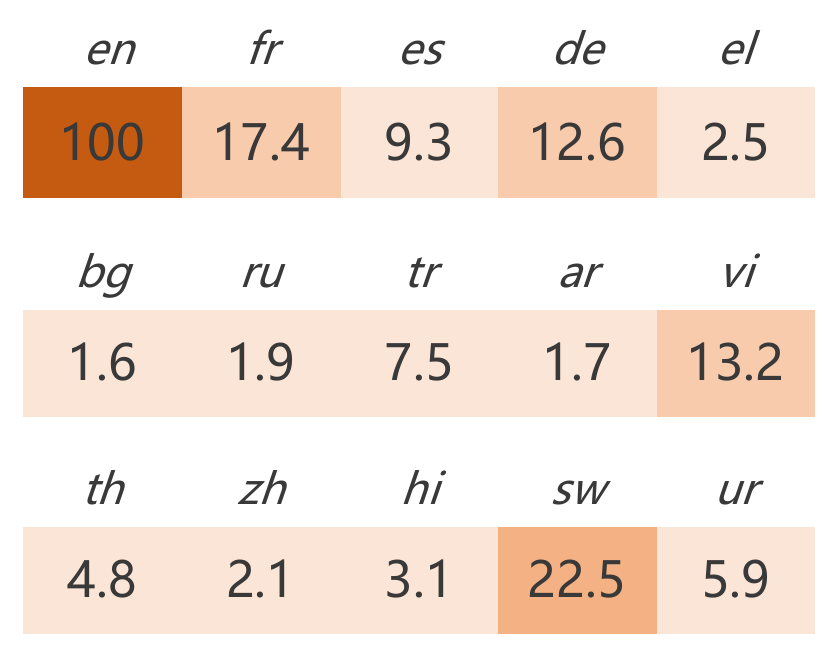}
        \caption{XNLI}
        \label{fig:xnli_sim}
    \end{subfigure}
\caption{
\label{fig:sda_xnli_sim}
The domain similarity of the SDA and XNLI dataset. (a) SDA: each element of the matrix represents the vocabulary overlap of source (rows) and target domain (columns). (b) XNLI: the similarity between the English domain and all other domains. 
}
\end{figure}

\paragraph{Domain Fusion with Different Similarities:}
In this part, we study the effect of domain fusion training with different domain similarities.
We first analyzed the similarities between each domain using the vocabulary overlaps \citep{gururangan-etal-2020-dont}. Specifically, we first build the vocabulary on the corpus of each domain. Then for any two domains, we calculate the overlap of their top 10k frequent vocabulary words as their domain similarity. The domain similarities of the SDA and XNLI dataset are separately shown in \ref{fig:sda_sim} and \ref{fig:xnli_sim}. The results confirm our intuition that the gaps between each domain in XNLI are much larger than SDA. This conclusion makes sense because all texts in UDA are English Amazon reviews, while texts in XNLI are in different languages.

\begin{figure*}[ht]
\centering
\includegraphics[width=395px]{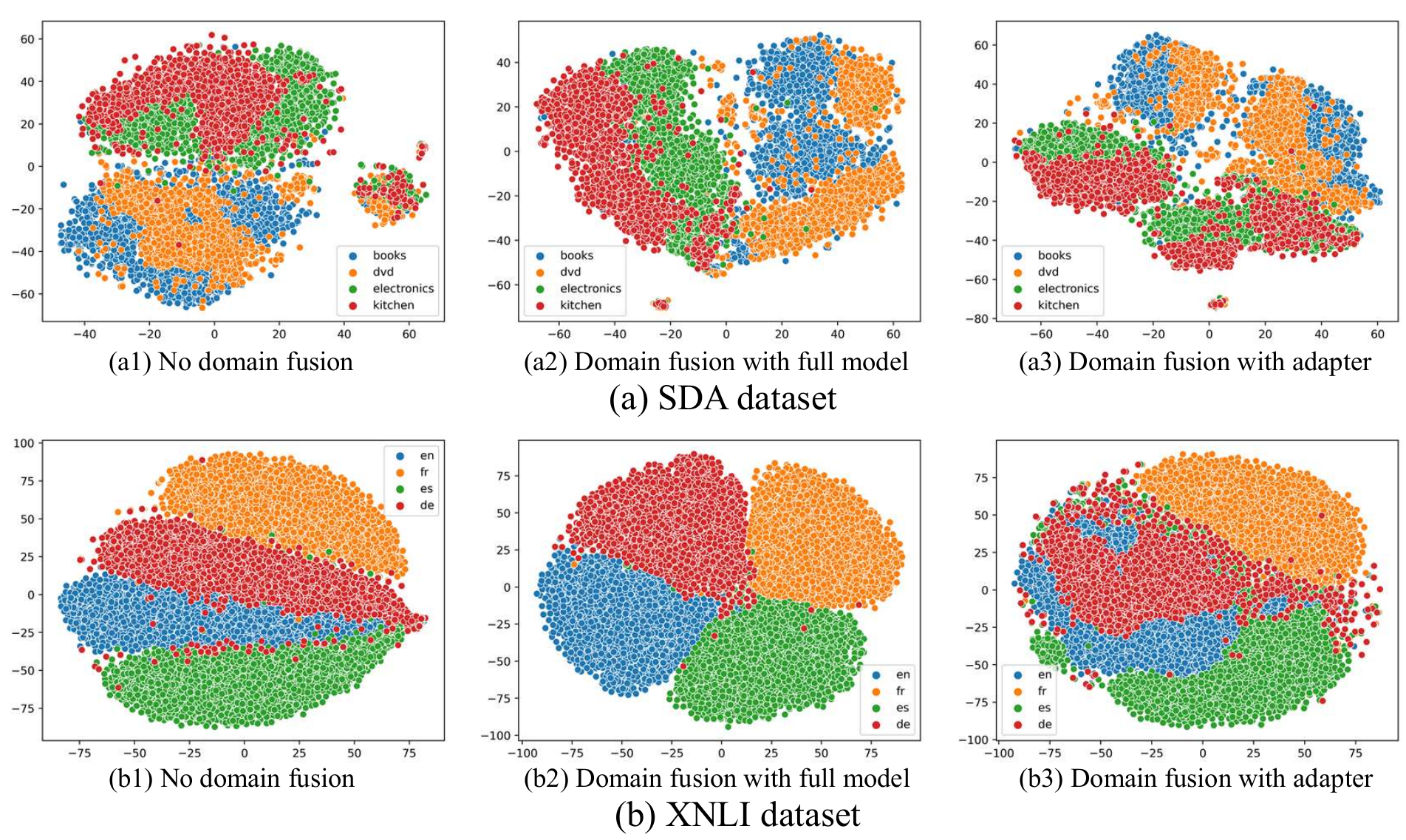}
\caption{Hidden representation of different domain fusion (DF) ways. For both SDA and XNLI (only show four languages for clarity), we plot the figures for three cases, (1) No domain fusion: the original pretrained model without DF. (2) Domain fusion with full model: perform DF on the original pretrained model. (3) Domain fusion with adapter: perform DF on the adapter-based model.}
\label{fig:hidden}
\end{figure*}

The results in Figure \ref{fig:avg_results} suggest that our domain-fusion training step is more effective at improving the UDA performance when the data used in this step come from similar domains. Specifically, the domain-fusion process brings an absolute gain of 1.14\% on the SDA dataset (i.e., from 92.17\% of Ada-FT to 93.31\% of Ada-TSA). In contrast, this gain drops to 0.22\% on the XNLI dataset (i.e., from 72.49\% of Ada-FT to 72.71\% of Ada-TSA), which has larger domain gaps. 
To get a deeper insight into different ways of domain fusion training, we compute their hidden representation of the last layer and reduce the dimension using the t-SNE algorithm \citep{van2008visualizing}, which are shown in Figure \ref{fig:hidden}. We can see that the dataset with smaller domain similarity (XNLI) brings less changes in hidden representation than the one with larger domain similarity (SDA). And it is more effective to perform the domain fusion training with the adapter module than with the full model.


\section{Conclusion}

This paper presents an adapter-based fine-tuning approach for unsupervised domain adaptation. Trainable adapter modules are inserted in a pre-trained LM, and a two-step training process is introduced to learn the parameters of these adapters. We demonstrate our method on two datasets with different sizes and domain similarities. The results show that the introduced adapter modules facilitate the adaption process, and our two-step training approach helps to further improve the UDA performance. As future works, we will explore more effective architectures of the adapter module.

\setcitestyle{numbers}
\bibliography{custom}
\bibliographystyle{abbrvnat}

\appendix

\section{Model and Training Details}
\label{sec:training_details}
In this appendix, we provide more details about the model and training configuration.

The hyper-parameter settings of different models and fine-tuning process on the SDA and XNLI dataset are provided in Table \ref{tab:model_training_details}. In the domain-fusion training process, the corpus used to optimize the MLM loss contains data sampled from the source domain and all target domains. The learning rate of domain-fusion training is set to 2e-5 for the model with full-parameter update (i.e., Full-TSA) and 5e-5 for the model with adapter-parameter update (i.e., Ada-TSA). Additionally, the training of all models starts with a warm-up step of 1K, and the learning rate decreases linearly with the number of training steps. The Adam optimizer is used with the parameter $\beta_1=0.9$, $\beta_2=0.999$ and $\epsilon=10^{-8}$. All the models are trained on the GTX2080Ti GPU.

\begin{table*}[ht]
\centering
\begin{tabular}{c|cccccccc}
\toprule
Dataset & Model & AdaS & LR & BZ & TrE & ToP & TrP & MaxL\\
\midrule
\multirow{4}*{SDA} & Full-FT & - & 2e-5 & 16 & 10 & 125M & 125M & 512 \\
 & Full-TSA & - & 2e-5 & 16 & 10 & 125M & 125M & 512 \\
 & Ada-FT & 128 & 5e-5 & 16 & 10 & 127M & 2.9M & 512 \\
 & Ada-TSA & 128 & 5e-5 & 16 & 10 & 127M & 2.9M & 512 \\
\midrule
\multirow{4}*{XNLI} & Full-FT & - & 2e-5 & 32 & 10 & 278M & 278M & 128 \\
 & Full-TSA & - & 2e-5 & 32 & 10 & 278M & 278M & 128 \\
 & Ada-FT & 256 & 5e-5 & 32 & 10 & 283M & 5.3M & 128 \\
 & Ada-TSA & 256 & 5e-5 & 32 & 10 & 283M & 5.3M & 128 \\
\bottomrule
\end{tabular}
\caption{Detailed configurations of the models and training process. Abbreviation description: LR: learning rate, BZ: batch size, TE: training epoch, ToP: total parameter, TrP: trainable parameter, AdaS: adapter size, MaxL: max sequence length.}
\label{tab:model_training_details}
\end{table*}

\section{Supplementary Experiments}
\label{sec:sup_exp}
This appendix provides the experiment results on a larger multi-domain sentiment analysis dataset to support the analysis in our paper. Specifically, this dataset is collected from Kaggle\footnote{https://www.kaggle.com}, and it contains reviews from multiple sources (i.e., domains) including Amazon ($\mathcal{A}$), IMDB ($\mathcal{I}$), TripAdvisor ($\mathcal{T}$) and Yelp ($\mathcal{Y}$). We denote this dataset as \textbf{MSS} (\underline{M}ulti-\underline{S}ource \underline{S}entiment dataset).
The statistic of MSS is summarized in Table \ref{tab:mss_datasize} and the experimental results are shown in Table \ref{tab:multi_source_result}.

\begin{table*}[ht]
\centering
\begin{tabular}{c|ccc}
\toprule
Source & Train & Dev & Test\\
\midrule
Amazon & 20K & 5K & 5K \\
IMDB & 20K & 5K & 5K \\
TripAdvisor & 8K & 1K & 1K \\
Yelp & 20K & 5K & 5K \\
\bottomrule
\end{tabular}
\caption{Statistics of the MSS dataset. Note that the Train, Dev, and Test set are all balanced.}
\label{tab:mss_datasize}
\end{table*}

\begin{table*}[!tbh]
\centering
\setlength\tabcolsep{7pt}
\begin{tabular}{c|ccc|c}
\toprule
$src \rightarrow tgt$ & Full-FT & Full-TSA & Ada-FT & Ada-TSA \\
\midrule 
$\mathcal{A} \rightarrow \mathcal{I}$ &  93.30 \small $\pm$ 0.16  & 93.54 \small $\pm$ 0.15 & \textbf{93.79} \small$\pm$ 0.24 & 93.72 \small$\pm$ 0.23 \\
$\mathcal{A} \rightarrow \mathcal{T}$ &  82.72 \small $\pm$ 1.51  & \textbf{86.58} \small $\pm$ 0.65 & 86.22 \small$\pm$ 1.79 & 86.10 \small$\pm$ 1.52 \\
$\mathcal{A} \rightarrow \mathcal{Y}$ &  95.43 \small $\pm$ 0.51  & 96.14 \small $\pm$ 0.27 & \textbf{96.25} \small$\pm$ 0.15 & 95.88 \small$\pm$ 0.42 \\
\midrule 
$\mathcal{I} \rightarrow \mathcal{A}$ &  90.62 \small $\pm$ 1.30  & 89.24 \small $\pm$ 1.66 & 92.60 \small$\pm$ 0.27 & \textbf{93.24} \small$\pm$ 0.37 \\
$\mathcal{I} \rightarrow \mathcal{T}$ &  85.10 \small $\pm$ 1.50  & 82.28 \small $\pm$ 2.37 & 86.66 \small$\pm$ 0.15 & \textbf{89.30} \small$\pm$ 0.63 \\
$\mathcal{I} \rightarrow \mathcal{Y}$ &  92.82 \small $\pm$ 0.90  & 92.59 \small $\pm$ 1.03 & 95.19 \small$\pm$ 0.14 & \textbf{95.48} \small$\pm$ 0.09 \\
\midrule
$\mathcal{T} \rightarrow \mathcal{A}$ &  92.47 \small $\pm$ 0.26  & \textbf{93.71} \small $\pm$ 0.12 & 92.13 \small$\pm$ 0.30 & 92.10 \small$\pm$ 0.29 \\
$\mathcal{T} \rightarrow \mathcal{I}$ &  88.64 \small $\pm$ 0.87  & \textbf{89.64} \small $\pm$ 0.06 & 87.60 \small$\pm$ 0.81 & 88.81 \small$\pm$ 0.33 \\
$\mathcal{T} \rightarrow \mathcal{Y}$ &  93.15 \small $\pm$ 0.40  & \textbf{94.58} \small $\pm$ 0.09 & 94.57 \small$\pm$ 0.12 & 93.53 \small$\pm$ 0.39 \\
\midrule
$\mathcal{Y} \rightarrow \mathcal{A}$ &  94.50 \small $\pm$ 0.30  & 95.16 \small $\pm$ 0.07 & 95.00 \small$\pm$ 0.12 & \textbf{95.40} \small$\pm$ 0.10 \\
$\mathcal{Y} \rightarrow \mathcal{I}$ &  91.99 \small $\pm$ 0.07 & \textbf{93.24} \small $\pm$ 0.19 & 92.49 \small$\pm$ 0.08 & 92.82 \small$\pm$ 0.11 \\
$\mathcal{Y} \rightarrow \mathcal{T}$ &  86.36 \small $\pm$ 1.02  & 89.16 \small $\pm$ 0.54 & 85.20 \small$\pm$ 0.32 & \textbf{89.24} \small$\pm$ 0.53 \\
\midrule
Avg. &  90.59  & 91.32 & 91.47 & \textbf{92.13} \\
\bottomrule
\end{tabular}
\caption{The accuracy on the MSS dataset and standard deviations of five independent runs.}
\label{tab:multi_source_result}
\end{table*}

\paragraph{Discussion:}
Experiments on the MSS dataset corroborate the discussion about the effect of data size and domain similarities in Section 4.3 of the main text:

1) Comparing the results of SDA and MSS, which have the same task (i.e., sentiment analysis) but different data sizes, a larger gain is obtained with the injection of the adapter module when the data size is larger. Specifically, the adapter module improves the average accuracy from 90.59\% of Full-FT to 91.47\% of Ada-FT on the MSS dataset (an absolute improvement of 0.88\%) while only 92.10\% of Full-FT to 92.17\% of Ada-FT on the SDA dataset (an absolute improvement of 0.07\%). This proves that more data facilitates to learn the parameters of the adapter module from random initialization.

2) Comparing the results of MSS and XNLI, which both have enough corpus for their corresponding task, the domain-fusion training is more effective on the MSS dataset than on the XNLI dataset. Specifically, the averaged accuracy improves from 91.47\% of Ada-FT to 92.13\% of Ada-TSA (the gain is 0.66\%) on the MSS dataset. However, this performance improvement drops to 0.22\% on the XNLI dataset. (from 72.49\% of Ada-FT to 72.71\% of Ada-TSA). This proves that the domain-fusion training brings a larger gain when the source and target domains are more similar.

\section{Link of Our Code}
\label{sec:link}
Our experiments are implemented utilizing the Transformers\footnote{https://github.com/huggingface/transformers} framework of Huggingface. To support the reproducibility of our work, we provide an anonymous download link \footnote{https://drive.google.com/file/d/1fHl4w1DmbRDon-3OxWsYz6AeQ6FmweWP/view?usp=sharing} of our code.

\end{document}